\documentclass[conference]{IEEEtran}
\usepackage{amsmath,amsthm,amssymb,amsfonts}
\usepackage{algorithmic}
\usepackage{graphicx}
\usepackage{textcomp}
\usepackage{xcolor}
\usepackage{tikz}
\usetikzlibrary[datavisualization]
\usepackage{comment}
\usepackage{gensymb}
\usetikzlibrary{calc}
\usetikzlibrary{decorations.pathreplacing} 
\usepackage{pgfplots}
\usepackage{pgfplotstable}
\usepgfplotslibrary{fillbetween}
\usepackage[normalem]{ulem}
\graphicspath{{images/}}
\usepackage{mgraphics}
\usepackage[flushleft]{threeparttable}

\usepackage[top=0.72in, bottom=1.0in, left=0.63in, right=0.66in]{geometry} 
\usepackage[subrefformat=parens,labelformat=parens,caption=false,font=footnotesize]{subfig}

\def\BibTeX{{\rm B\kern-.05em{\sc i\kern-.025em b}\kern-.08em T\kern-.1667em\lower.7ex\hbox{E}\kern-.125emX}}

\newcommand{\parenthesis}[1]{\left(#1\right)}
\def\params{\mathrmbold{w}}
\def\obsLocal{\mathrmbold{o}_j^{\textsc{L}}(t)}
\def\obsGlobal{\mathrmbold{o}_j^{\textsc{G}}(t)}

\newcommand\blfootnote[1]{%
  \begingroup
  \renewcommand\thefootnote{}\footnote{#1}%
  \addtocounter{footnote}{-1}%
  \endgroup
}

\def\tablescale{1}

\newcommand{\titleheader}{This work has been accepted for publication in IEEE PIMRC 2021}

\makeatletter
\def\ps@headings{%
\def\@oddhead{\mbox{}\scriptsize \titleheader}
\def\@oddfoot{\scriptsize \@date \hfil }%
}

\def\ps@IEEEtitlepagestyle{%
\def\@oddhead{\mbox{}\scriptsize \titleheader \rightmark \hfil }%
\def\@oddfoot{\scriptsize \@date \hfil }%
}
\makeatother

\title{Transferable and Distributed User Association Policies for 5G and Beyond Networks}

\author{\IEEEauthorblockN{Mohamed Sana$^{1}$, Nicola di Pietro$^{2}$, Emilio Calvanese Strinati$^{1}$}
\IEEEauthorblockA{$^{1}$CEA-Leti, Université Grenoble Alpes, F-38000 Grenoble, France\\
$^{2}$Athonet, via Cà del Luogo 6/8, 36050, Bolzano Vicentino (VI), Italy\\
Email: \{mohamed.sana, emilio.calvanese-strinati\}@cea.fr, nicola.dipietro@athonet.com}}

\begin{document}
\maketitle






\begin{abstract}

We study the problem of user association, namely finding the optimal assignment of user equipment to base stations to achieve a targeted network performance. In this paper, we focus on the \emph{knowledge transferability} of association policies. Indeed, traditional non-trivial user association schemes are often scenario-specific or deployment-specific and require a policy re-design or re-learning when the number or the position of the users change. In contrast, transferability allows to apply a single user association policy, devised for a specific scenario, to other distinct user deployments, without needing a substantial re-learning or re-design phase and considerably reducing its computational and management complexity. To achieve transferability, we first cast user association as a multi-agent reinforcement learning problem. Then, based on a neural \emph{attention mechanism} that we specifically conceived for this context, we propose a novel distributed policy network architecture, which is transferable among users with \emph{zero-shot generalization capability} i.e., without requiring additional training.Numerical results show the effectiveness of our solution in terms of overall network communication rate, outperforming centralized benchmarks even when the number of users doubles with respect to the initial training point.

\end{abstract}

\blfootnote{This work has been partially supported by the CPS4EU Project, Nr. 826276 and the H2020 RISE-6G Project.}

\vspace{-0.1cm}
\section{Introduction}
With the proliferation of smart connected devices, the cyber and physical spaces are fusing, turning humans, objects and events more and more into exponentially growing sources of digital information \cite{emilio2019}. 
To cope with this, modern wireless networks, such as 5G networks, are becoming denser and heterogeneous with the coexistence of base stations (BSs) operating at different frequencies. In this context, \emph{user association}, namely efficiently finding optimal assignments of user equipments (UEs) to BSs to achieve a targeted network performance, is a crucial challenge because it directly affects the network spectral efficiency and the users' perceived quality of service. In general, for dense networks, this is a challenging task as it involves non-convex and combinatorial optimization problems, whose complexity grows exponentially with the number of UEs. This difficulty is even exacerbated in highly dynamic networks such as in millimeter-wave (mmWave) networks, subject to frequent changes of the radio environment due to highly directional transmissions and variable channel conditions \cite{DeDomenico2017}. 

To address these issues, we propose a scalable and easily manageable user association policy. 
Our proposed approach is conceived with a specific focus on the key aspect of 
\emph{transferability}, which 
allows to apply a user association strategy or policy acquired in a specific scenario (\eg a network deployment) to distinct but related ones, without needing to substantially redesign, recompute or relearn it. 
This considerably reduces the computational complexity of user association during the network operations and makes the policy adapted to distributed and dynamic scenarios. So far, despite their many appreciable features, solutions of the literature lack 
transferability. In~\cite{athanasiou}, a distributed user association scheme is proposed using Lagrangian tools. The user association is reformulated as a non-cooperative game with local interactions in \cite{lui} and as a matching game in \cite{Alizadeh2019c}. However, every time the radio environment changes due to, for instance, the arrival or departure of UEs, this solution needs to be recomputed to seek a new convergence point and to correct a possible drift from optimality.
Recently, a deep neural network (NN) architecture was introduced in \cite{zhou} that predicts the user association and power allocation. Similarly, authors in \cite{RiuLiu2020} formulated the problem of user association as a multi-label classification problem. In \cite{zhao, sana2019UA, sana2020UA}, the authors proposed an approach based on distributed \emph{multi-agent reinforcement learning}. However, whenever the number of UEs change, these solutions either require a new learning procedure \cite{zhou, sana2020UA} or to entirely redesign and retrain the architecture of the NNs \cite{zhao}. With the complexity associated to the re-computation or re-learning procedures, such approaches are unsuitable to dynamic networks, characterized by a frequent change of the radio environment due to \eg mobility or the arrival and departure of UEs.

Here, to overcome the inadequacy of state-of-the-art solutions, we first cast the user association problem as a multi-agent reinforcement learning problem, aimed at optimizing predefined network utility functions common to all UEs. Then, by conveniently adapting to this context a \emph{neural attention mechanism}, we successfully design a global policy network architecture (PNA) that is transferable among UEs. 
Given this PNA, UEs learn a common association policy leveraging their local (and global, if available) observations. 
The learned policy has \emph{zero-shot generalization capability}, thus considerably reducing the computational complexity of the user association task. Indeed, thanks to the proposed architecture, a policy learned in a specific deployment can be transferred to another one without requiring substantial additional training procedure. Consequently, as desired, the proposed mechanism adapts well and by design to variations in the number of UEs or changes in the geometry of the network (namely the geographical positions of the UEs). Moreover, the proposed mechanism incorporates channels' and UEs' traffic dynamics during the training phase to foster better adaptability to the variations of radio channel quality and requested quality of service (QoS). Finally, our proposed solution can be implemented either in a centralized or in a distributed manner to trade-off computational complexity and/or signaling overhead. 

\section{System Model and Problem Formulation}
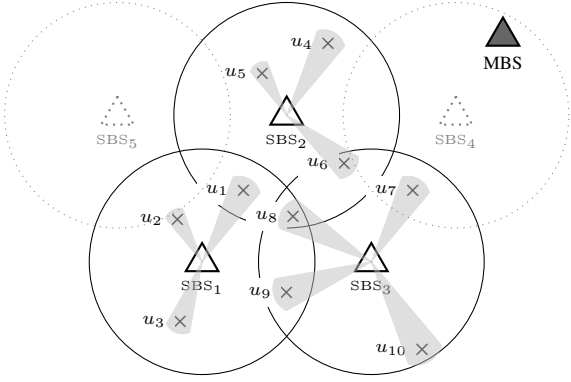
\begin{figure}
    \centering
    \begin{tikzpicture}[label distance=-2mm]
        \basestation{{\scriptsize MBS}}{fill=black!60, shift={(4, 3)}}{(0., 0)};
        
        \cell{1.5}{1}{shift={(0, 0)}}{(0,0)};
        \cell[(1.125, 1.95)]{1.5}{2}{}{(0,0)};
        \cell[(2.25, 0)]{1.5}{3}{}{(0,0)};

        \user{120}{0.75}{10}{gray!50}{2};
        \user{250}{0.95}{10}{gray!50}{3};
        
        \user[(1.125, 1.95)]{60}{1.2}{10}{gray!50}{4};
        \user[(1.125, 1.95)]{120}{0.75}{10}{gray!50}{5};
        \user[(1.125, 1.95)]{320}{1.1}{10}{gray!50}{6};
        
        \user[(2.25, 0)]{60}{1.2}{10}{gray!50}{7};
        \user[(2.25, 0)]{150}{1.3}{10}{gray!50}{8};
        \user{60}{1.2}{10}{gray!50}{1};
        \user[(2.25, 0)]{200}{1.3}{10}{gray!50}{9};
        \user[(2.25, 0)]{300}{1.45}{10}{gray!50}{10};

        
        \cell[(3.375, 1.95)]{1.5}{4}{gray, dotted}{(0,0)};
        \cell[(-1.125, 1.95)]{1.5}{5}{gray, dotted}{(0,0)};        

    \end{tikzpicture}
    \caption{Network topology for $N_s=3$ SBSs, 1 MBS, and $K=10$ UEs.}
    \label{fig:sys-model}
\end{figure}

\subsection{Network Model}\label{systemmodel}

We consider the system model of \fig{fig:sys-model} as in \cite{sana2020UA}: we focus on downlink communications in a network of $K(t)$ UEs located at time $t$ in a region of the bi-dimensional Euclidean space, covered by $N_s$ mmWave small base stations (SBSs) and a sub-6 GHz macro base station (MBS), to enable ubiquitous network coverage. Let $\mathcal{A} = \{0,1,\dots, N_s\}$ be the set of BSs, where $0$ indexes the MBS, and $\mathcal{U}(t) = \{1,2, \dots, K(t)\}$ be the set of UEs in the network. We call a network deployment $\mathcal{D}(t)$, a collection of positions of all UEs in the network:
\begin{equation}
    \mathcal{D}(t) = \left\{ \left( x_j(t), y_j(t) \right), j\in\mathcal{U}(t)\right\},
\end{equation}
where $x_j(t)$ and $y_j(t)$ denote respectively the 
two coordinates of UE $j$ in deployment $\mathcal{D}(t)$, expressed with respect to a reference system common to all UEs and BSs.

We denote with $\mathcal{A}_j(t) = \{i, d_{i,j}(t) \leq R_0\} \cup \{0\} \subseteq \mathcal{A}$ the subset of BSs that can be associated with UE $j$ at time $t$. Here, $R_0$ is the SBSs' coverage range and $d_{i,j}(t)$ is the distance from UE $j$ to SBS $i$, which we assume to be able to support at most $N_i$ UEs simultaneously. Also, we assume that a UE is associated to exactly one BS at a time and each BS $i$ communicates to its served UEs using equal transmit power. We adopt the Friis propagation loss model \cite{bai2015coverage}, according to which the power received by a UE, $P^{\mathrm{Rx}}$, is given as a function of the distance $d$ between the UE and its serving BS:
\begin{equation}\label{eq:channel-model}
    P^{\mathrm{Rx}}(d) = hP_s^{\mathrm{Tx}} G_s^{\mathrm{Tx}} G_s^{\mathrm{Rx}} C_s d^{-\eta_s}, ~~ s\in\{\mathrm{MBS}, \mathrm{SBS}\}.
\end{equation}
Here, $C_s$ denotes the path-loss constant, $\eta_s$ is the path-loss exponent, $P_s^{\mathrm{Tx}}$ is the transmit power \wrt BS $s$ and $h$ denotes the fading coefficient. We assume $m$-Nakagami fading for UE-SBS links whereas UE-MBS links experience Rayleigh fading, which is a special case of $m$-Nakagami, where $m=1$. The gains of the transmitter and receiver antennas \wrt BS $s$ are $G_s^{\mathrm{Tx}}$ and $G_s^{\mathrm{Rx}}$ respectively.
In addition, we assume that mmWave SBSs allocate all the available band to their served UEs, whereas the MBS equally shares its band across its UEs. Also, we assume that UEs and BSs perform beam steering and training in advance and ignore their impact when optimizing user association. Finally, we assume that there exists a central controller, collocated with the MBS, able to collect and forward information to the UEs.

\subsection{Problem formulation}

At time $t$, each UE $j$ requests a data rate $D_j(t)$ from its serving BS $i$ to satisfy a certain QoS. It then experiences a signal-to-interference plus noise ratio $\mathrm{SINR}_{i,j}$, which comprises both intra-cell and inter-cell interference. We say that UE $j$'s QoS is fully satisfied at time $t$, if the achievable data rate ${R_{i,j}(t) = B_{i,j}\mathrm{log}_2\parenthesis{1+\mathrm{SINR}_{i,j}(t)}}$, given by the Shannon capacity of the channel between UE $j$ and BS $i$, is greater than $D_j(t)$. Therefore, the effective communication rate between UE $j$ and its serving BS equals $\mathrm{min}\parenthesis{R_{i,j}(t), {D_{j}(t)}}$. Hence, we define an $\alpha$-fair \emph{network utility function}~\cite{Srikant2014} as follows: 
\begin{align}\label{def-rate}
R(t) &=\sum_{i\in \mathcal{A}}\sum_{j \in \mathcal{U}(t)}x_{i,j}(t)U_{\alpha}\parenthesis{\mathrm{min}\parenthesis{R_{i,j}(t), D_{j}(t)}},\\\nonumber
    &=\sum_{i\in \mathcal{A}}\sum_{j \in \mathcal{U}(t)}x_{i,j}(t)U_{\alpha}\parenthesis{\kappa_{i,j}(t)D_{j}(t)},
\end{align}
where $x_{i,j}(t)=1$ indicates that UE $j$ is associated with BS $i$ at time $t$; otherwise $x_{i,j}(t)=0$ and $\kappa_{i,j}(t) = \mathrm{min}\parenthesis{1, \frac{R_{i,j}(t)}{D_{j}(t)}} \in [0,1]$ indicates the QoS satisfaction of UE $j$ \wrt its associated BS $i$, which is fully satisfied when $\kappa_{i,j}(t)=1$. $U_{\alpha}(\cdot)$ is the $\alpha$-fair utility function given in \cite[Section 2.2]{Srikant2014} as follows:
\begin{equation}
    U_{\alpha}(x) ={} \left \{
        				\begin{array}{l l}
        					(1-\alpha)^{-1}{x^{1-\alpha}}, &\text{~if~} \alpha \geq 0 \text{~and~} \alpha \neq 1,\\
        					\mathrm{log}(1+x), &~\alpha =1.
        				\end{array}
			        \right.
\end{equation}

\textcolor{black}{Given a network deployment $\mathcal{D}(t)$}, we formulate the user association problem to maximize 
\eqref{def-rate}  at time $t$ as follows:
\begin{IEEEeqnarray}{lCl}\label{eq:UApb}
&\underset{\{x_{i,j}(t)\}}{\mathrm{maximize}}~& R(t), \IEEEyesnumber\IEEEyessubnumber\label{eq:Obj}\\
&\mathrm{subject~to~}& x_{i,j}(t) \in \{0,1\}, ~ \forall i,j,\IEEEyessubnumber\label{eq:C1}\\
	 &{}&\sum_{j \in \mathcal{U}(t)}x_{i,j}(t) \leq N_{i}, ~ \forall i \in \mathcal{A}\backslash \{0\},\IEEEyessubnumber\label{eq:C2}\\
	 &{}&\sum_{i \in \mathcal{A}_j(t)}x_{i,j}(t) = 1, ~ \forall j \in \mathcal{U}(t).\IEEEyessubnumber\label{eq:C3}
\end{IEEEeqnarray}
Constraint (\ref{eq:C1}) indicates that the $x_{i,j}(t)$ are binary variables. The number of resources available at each SBS is limited; this is highlighted in \eqref{eq:C2}, by constraining the number of UEs simultaneously associated with a given SBS $i$, to be lower than $N_{i}$. Finally, \eqref{eq:C3} ensures that, in our setting, each UE is associated with exactly one BS. Depending on $\alpha$, problem \eqref{eq:UApb} guarantees different fairness criteria in the user association.
In particular, we will focus on $\alpha=0$ and $\alpha=1$, corresponding respectively to sum-rate maximization and proportional fairness, widely used in the literature \cite{LiuUESurvey2016}.

\textcolor{black}{
Although \eqref{eq:UApb} appears as a standard user association problem, solutions of the literature are often scenario-specific or deployment-specific. In other words, they assume either a pre-sized or a fixed set of static UEs. Here, we are interested in a different approach: first, we are looking for a policy that can be applied at each time $t$ by a user, based on its instantaneous observations of the environment. Then, taking into account the targeted optimization objective, we are interested in an association policy, which, once learned, is also transferable and capable of solving problem \eqref{eq:UApb} at each time $t$ regardless of the location and the number of UEs in the network, without the need of being relearned. This policy must be able to adapt to the departure or arrival of UEs from and in the network, as both events have an impact on the optimal user association. Also, a policy learned in a scenario of $K_1$ UEs has to be effectively applicable to a scenario of $K_2 \neq K_1$ UEs without additional training. For this purpose, the architecture of the association policy needs to be transferable, as well as the learned policy.}


\section{Proposed Transferable User Association policy}

\subsection{On transferable policy network architecture}
\label{sec:PNA_design}
\begin{figure}
    \centering
    \includegraphics[scale=0.71]{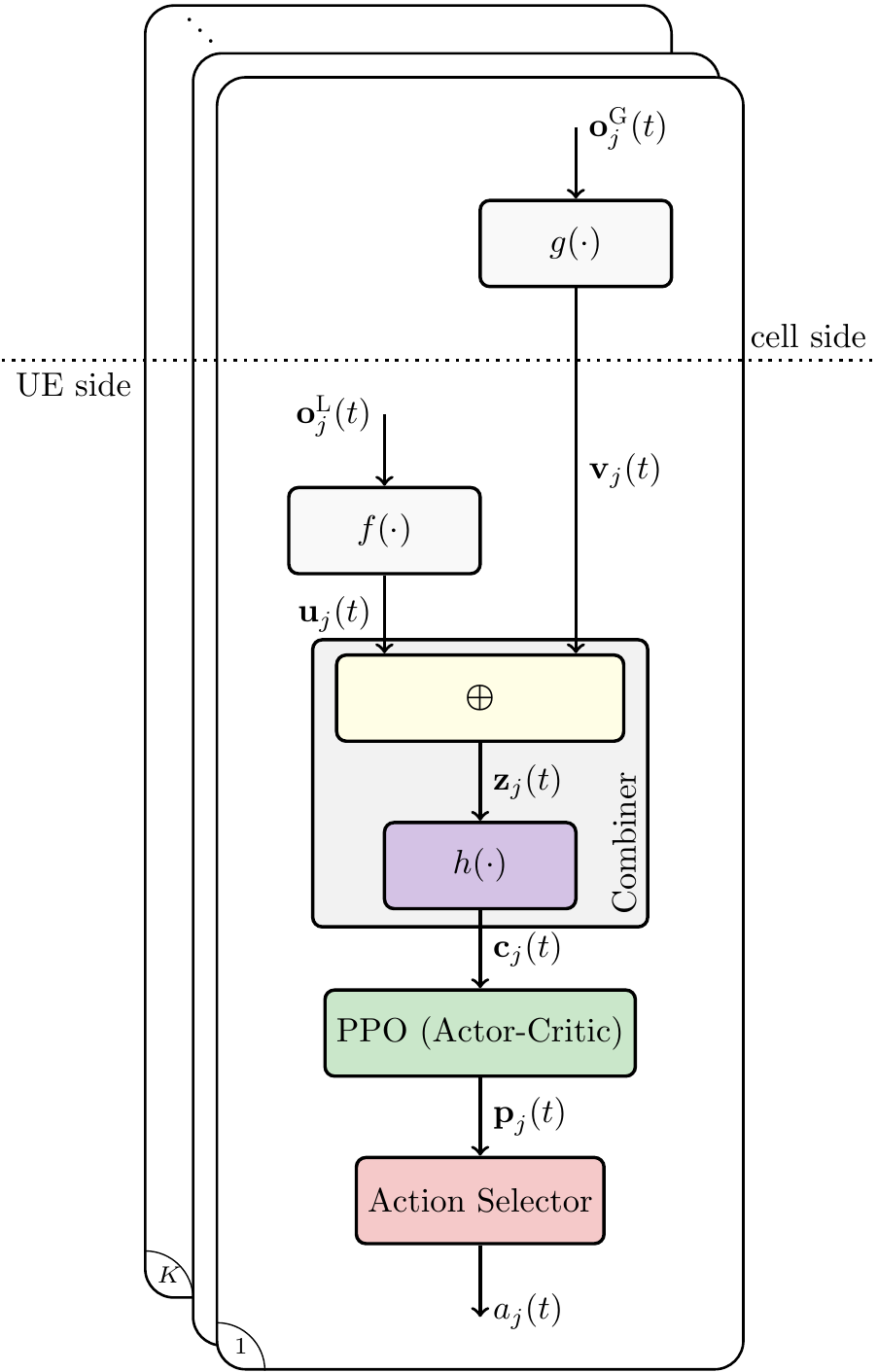}
    \caption{UE association policy network architecture, shared across all UEs.}
    \label{fig:PNA}
\end{figure}

In order for the policy architecture to be transferable, an adequate design of the PNA components is required. Our objective, in fact, is to construct a policy architecture whose size does not vary with the number of UEs in the network, which is bound to change over time. In the following, we will describe the policy network architecture of \fig{fig:PNA}, which allows the transferability of the association policy.

\subsubsection{UE local observation encoding}
In this study, we assume that at each time $t$, each UE $j$ can estimate the received signal strength (RSS) and the corresponding angle of arrival (AoA) \wrt its surrounding BSs. We denote with $\mathrm{RSS}_{i,j}$ and $\vartheta_{i,j}$ the estimated RSS and AoA of UE $j$ \wrt BS $i$, respectively. Moreover, as in \cite{sana2020UA}, a UE receives an acknowledgment (ACK) signal whenever its connection request succeeds ($\mathrm{ACK}_j=1$) or is denied ($\mathrm{ACK}_j=0$), which may happen due to the limited resources available at each BS \eqref{eq:C2}; we call this event a \emph{collision}. When it happens, each BS selects among the colliding UEs the best ones to associate with, according to their association probability that we define later. Next, we define UE $j$ local state, $\obsLocal$, as follows:
\begin{align}
    \begin{split}
        \obsLocal {}={}& \bigg\{a_j(t-1), R_{a_j(t-1),j}, R(t-1), \\
        &{} \mathrm{ACK}_j, \left\{\mathrm{RSS}_{i,j}(t)\right\}_{i \in \mathcal{A}_j(t)}, \left\{\vartheta_{i,j}\right\}_{i \in \mathcal{A}_j(t)} \bigg\}.
    \end{split}
\end{align}
Here, $R_{a_j(t-1),j}$ represents the achievable rate when UE $j$ is associated with the BS indexed by $a_j(t-1)$. Note that the size of $\obsLocal$ does not depend on the number of UEs, in sharp contrast with \cite{zhao}. Then, we obtain the $n$-dimensional local encoding vector $\mathrmbold{u}_j(t) = f(\obsLocal)$, where $f: \mathbb{R}^l \rightarrow \mathbb{R}^n$ is a NN, and $l$ is the size of the vector obtained after the concatenation of the elements in $\obsLocal$.

\subsubsection{UE global observation encoding}
After taking an action $a_j(t)$, the controller can encode for UE $j$ some meaningful global state (i.e. macro) information $\obsGlobal$ such as the estimated position of UEs of interfering links, i.e., of active mmWave links, the load of each BS, etc. However, embedding more information does not necessarily imply performance improvement. Indeed, as the agent's state space also increases, more exploration is required to discover the intrinsic state/action relation at the risk of misleading the agent. In our scenario, we consider that the information about the actual rate perceived by each UE $j$ and the position of the potential interferers, i.e., the set of UEs $\mathcal{N}_j(t)$, susceptible to impact the association decision of UE $j$ through the interference resulting from their communications. Here, we consider $\mathcal{N}_j(t)$ as the $k$-nearest neighbors of UE $j$, whose size may vary with time, however solutions based on local interaction graphs can be considered \cite{lui}. Hence, we define $\obsGlobal$ as:
\begin{align}
    \begin{split}
        \obsGlobal {}={}& \bigg\{\varsigma_l = \left[x_l(t), y_l(t), R_{a_l(t-1),l}\right],~~l \in \mathcal{N}_j(t) \bigg\},
    \end{split}
\end{align}
from which, we construct UE $j$ global state encoding $\mathrmbold{v}_j(t)$. 

\vspace{0.1cm}
\noindent
\textbf{Fixed-size encoding.}
A  naive solution to construct $\mathrmbold{v}_j(t)$ is to first concatenate all elements in $\obsGlobal$ resulting in a vector of size $m(t)=3\times\card{\mathcal{N}_j(t)}$. Then, we obtain the local encoding vector $\mathrmbold{v}_j(t) = g(\obsGlobal)$, where $g: \mathbb{R}^m\rightarrow \mathbb{R}^n$ is also a NN. However, such an approach i) has limited scalability as the size of $\obsGlobal$, i.e., $m(t)$ varies with the number of UEs, especially in the neighborhood, and ii) requires ordering elements prior to concatenation, preventing from transferability. 


\vspace{0.1cm}
\noindent
\textbf{Order-agnostic and size-variable encoding.}
An efficient solution to the problem should be agnostic of the ordering in $\obsGlobal$. Moreover, in order to build a scalable and transferable architecture, the size of $\mathrmbold{v}_j$ should be independent of the length of $\obsGlobal$, thus, the size of UE $j$ neighborhood. To satisfy these properties, we adopt ideas from the \emph{dot-product attention mechanisms} in \cite{vaswani2017attention}. Hence, let $\mathrmbold{k}_j = g_{k}\parenthesis{\varsigma_j}$, $\mathrmbold{q}_j = g_{q}\parenthesis{\varsigma_j}$, and $\mathrmbold{\nu}_j = g_{\nu}\parenthesis{\varsigma_j}$, where {$g_{k}, ~g_{q}, ~g_{\nu}: \mathbb{R}^3 \rightarrow \mathbb{R}^n$} are also encoding functions (e.g., neural networks), and $\mathrmbold{k}_j$, $\mathrmbold{q}_j,$ $\mathrmbold{\nu}_j$ denote the \emph{key}, the \emph{query} and the \emph{value} associated with UE $j$, respectively. For a given UE $j$, we compute for each UE in its neighborhood $\mathcal{N}_j(t)$ a weight (or score) $\alpha_{k,j}$
\begin{equation}
    \alpha_{k,j} = \softmax\left(\left[\frac{\mathrmbold{q}_k \mathrmbold{k}_j^T}{\sqrt{n}}\right]_{k\in\mathcal{N}_j(t)}\right).
\end{equation}
Here, $\softmax(\cdot)$ is the softmax function also known as the normalized exponential function. Let $\mathrmbold{\alpha}_j = \left[\alpha_{k,j}, {k\in\mathcal{N}_j}\right]$. The vector  $\mathrmbold{\alpha}_j$ represents the interaction of UE $j$ with its neighbors. Then, we compute the encoding $\mathrmbold{v}_j$ by aggregating all values' information from the neighborhood as follows:
\begin{equation}\label{eq:dot-att-value}
    \mathrmbold{v}_j =  \sum_{k\in\mathcal{N}_j(t)} \alpha_{k,j} \mathrmbold{\nu}_k.
\end{equation}
By construction, the size of $\mathrmbold{v}_j$ in \eqref{eq:dot-att-value} is invariable with the size of $\mathcal{N}_j(t)$. Only its value can change depending of the aggregated information. That is to say, whenever the number of UEs varies, there is no need to change the PNA.

\subsubsection{Local and global information combining}
Now, once we obtain the UE local and global encoding vector, they are merged together to build its context understanding vector $\mathrmbold{c}_j(t) = h\parenthesis{\mathrmbold{z}_j(t)}$, i.e., its perception of the radio environment, where $\mathrmbold{z}_j(t) = \mathrmbold{u}_j(t) \oplus \mathrmbold{v}_j(t)$, with $\oplus$ denoting concatenation operation and $h:\mathbb{R}^{2n} \rightarrow \mathbb{R}^n$ is also taken here, as a NN.

Now, given the context vector $\mathrmbold{c}_j(t)$, the goal of the learning agent $j$ at each instant $t$, is to define an \textit{association probability vector} $\mathrmbold{p}_j(t) = [p_{0,j}, \ldots, p_{N_s,j}] \in [0,1]^{N_s+1}$ with $\sum_{i\in\mathcal{A}} p_{i,j} = 1$ and $p_{i,j} = 0$ for all $i \not \in \mathcal{A}_j(t)$. Then, the UE's action $a_j(t)$, which corresponds to a connection request towards the BS indexed by $a_j(t)$ in $\mathcal{A}_j(t)$, is sampled from the distribution characterized by the $p_{i,j}$. Thus, the learning problem  consists in deriving an association policy that optimizes $\mathrmbold{p}_j(t)$, such that sampling from it maximizes \eqref{def-rate}. 

Note that in this architecture, UEs' agents share the same model, i.e., $f(\cdot)$, $g(\cdot)$, and $h(\cdot)$ are common to all UEs. This setting does not preclude UEs from taking different actions as they do not observe the same inputs. In contrast, sharing the parameters among UEs enables a better skill transfer since there is only a unique policy (in contrast to having one policy per UE as in \cite{zhao}), which can be efficiently and simultaneously trained with the experiences of all UEs in a MARL framework using \textit{proximal policy optimization} (PPO) \cite{schulman2017ppo}.

\subsection{Proximal policy optimization}
In a MARL system, agents learn by interacting with a shared environment by making decisions following a Markov Decision Process (MDP). In MDP, the action $a_j(t)$ of an agent $j$ in a given state $\mathrmbold{s}_j(t)$ leads it to the next state $\mathrmbold{s}_j(t+1)$ and results in a reward $r_j(t)$. From the underlying \emph{experience} $e_j(t) = \{\mathrmbold{s}_j(t), a_j(t), r_j(t), \mathrmbold{s}_j(t+1)\}$, the agent learns its policy $\pi_{j,\params}(\cdot|\cdot)$, parameterized by $\params$, the set of PNA parameters, where $\pi_{j,\params}(a_j|\mathrmbold{s}_j)=p_{a_j(t), j}$ is the probability that agent $j$ takes action (or requests connection) $a_j$ in state $\mathrmbold{s}_j$, to maximize an accumulated long-term $\gamma$-discounted reward $G_j(t)=\sum_{\tau=t+1}^{T_e} \gamma^{\tau-t-1}r_j(\tau)$ over an \textit{episode} - a new network deployment - of duration $T_e$:
$\pi_{j,\params}^{*} = \argmax{\pi_j}{\mathbb{E}_t\left[G_j(t)\right]}$.
In our study, we consider the particular case of \emph{cooperative} MARL \cite{bucsoniu2010multi}, i.e., UEs share the same reward, hence, they are assigned to the same objective of maximizing the network utility function: $r_j(t) = R(t), ~\forall j$. Moreover, UEs also share the same policy parameters, i.e., $\pi_{j,\params} = \pi_{\params}, ~\forall j$. 

In general MARL, an agent has only access to a partial observation $\mathrmbold{o}_j(t) = \left\{\obsLocal, \obsGlobal\right\}$ of the actual state $\mathrmbold{s}_j(t)$, which is unknown, resulting in partially observable MDP \cite{omidshafiei2017}. Moreover, MARL is subject to non-stationarities due to simultaneous interactions of agents with the environment, which make the learning process more complex.
In the literature, \textit{policy gradient} algorithms are used to solve this problem \cite{Sutton1998}. We use an actor-critic mechanism to iteratively update the policy parameters $\params$ to minimize the $\epsilon$-clipped proximal loss:
\begin{equation}
\label{lossfunc}
\mathcal{L}(\mathrmbold{\params}){}={} \mathbb{E}_{\pi}\left[\mathrm{min}\left(\zeta(\mathrmbold{\params})\hat{A}, \mathrmbold{clip}\left(\zeta(\mathrmbold{\params}), 1-\epsilon_1, 1+\epsilon_2\right)\hat{A}\right)\right],
\end{equation}
where $\mathrmbold{clip}(x, a, b) = \mathrm{min}\left(\mathrm{max}\left(x, a \right), b\right)$, $\hat{A}(a_j,\mathrmbold{o}_j)$ denotes the advantage estimator
, which measures the advantage of selecting a given action in a given state, that we estimate using one step Temporal Difference error \cite{Sutton1998}. $\zeta(\mathrmbold{\params}) = \frac{\pi_{\mathrmbold{\params}}(a_j|\mathrmbold{o}_j)}{\pi_{\mathrmbold{\params}_{\rm old}}(a_j|\mathrmbold{o}_j)}$ is the probability ratio between current and previous update. By introducing the clipping effect, PPO pessimistically ignores updates (possibly destructive) that will lead to high changes between policy updates.\\
\noindent
\textbf{Hysteretic PPO.} Note that in vanilla PPO, $\epsilon_1=\epsilon_2$. However, in multi-agent environments, an agent should not be pessimistic in the same way for both \textit{positive} ($\zeta(\mathrmbold{\params})>1$) and \textit{negative} ($\zeta(\mathrmbold{\params})<1$) updates. In fact, due to the interaction of multiple agents with the environment and the common reward of the cooperative framework, an agent may receive a lower reward because of the bad behavior of its teammates. This may cause the user to change its policy at the risk to misleading it. To overcome this issue, following the concept of hysteretic Q-learning in \cite{matignon2007hysteretic}, we introduce \textit{hysteretic proximal policy optimization}, where we use $\epsilon_1$ and $\epsilon_2$ for negative and positive updates, with $\epsilon_1 < \epsilon_2$. In this way, an agent gives more importance to updates that improve its policy rather than to ones that worsen it. This setting is particularly important when agents do not have equal contribution to the team's reward and for decentralized learning. 

Note that the association policy can be efficiently trained in a centralized way with the experience of all agents or in a decentralized way, by leveraging approaches presented in \cite{Wijmans2020DDPPOLN}. 

To further make learning robust against the variability of the number of UEs over time, we introduce a \textit{UE dropout mechanism} with rate $p_0$, corresponding to the Bernoulli probability of a UE to be masked out in a given training episode, thus, appearing as non-existent in the cell for the others UEs.

\noindent
\textbf{On complexity.} 
In contrast to previous work where each UE learns its own specific policy without transferability \cite{sana2020UA}, here we have only one global policy that can be transferred to any UE in the network even new ones, thus considerably reducing the computation complexity. Also by using attention mechanism instead of Long Short Term Memories (LSTMs), we considerably reduce the PNA architecture. The counterpart is the aggregation of information in \eqref{eq:dot-att-value}. However, this process can be viewed as a message passing between UEs, where they only need to exchange their queries and values with BSs and only when there is a considerable change in the network (\eg arrival of new UEs) to limit signaling.

\section{Numerical Results}\label{res-section}
\begin{table}[!t]
    \centering
    \caption{Simulations parameters \protect\cite{sana2020UA}}
    \label{simu-params}
    \scalebox{\tablescale}{
    \renewcommand{\arraystretch}{1.1}
    \begin{threeparttable}
    \begin{tabular}{l||c|c}
       \hline
        & Macro cell & Small cell\\
       \hline
       \hline
        Parameters & \multicolumn{2}{c}{Values} \\
       \hline
        Carrier frequency $f_s$ & 2.0 GHz & 28 GHz\\
        \hline
        Path loss constant $C_s$ & \multicolumn{2}{c}{$\left(c/{4\pi f_s}\right)^2$, $c=3\times10^8 \mathrm{ms}^{-1}$}\\
        \hline
        Bandwidth & 10 MHz & 200 MHz\\
        \hline
        Thermal noise, $N_0$ & \multicolumn{2}{c}{-174 dBm/Hz}\\
        \hline
        Noise figure & 5 dB &0 dB \\
        \hline
        Shadowing variance & 9 dB & 12 dB\\
       \hline
        TX power, $P^{\mathrm{Tx}}$ & 46 dBm & 20 dBm\\
       \hline
        Antenna gain, $G_{\mathrm{max}}^{\mathrm{Tx}}$ / $G_{\mathrm{max}}^{\mathrm{Rx}}$ & 17 dBi / 0 dBi & \cite[diag. 2]{sana2020UA}\\ 
       \hline
        Radius, $R_0$ & & 50 m\\
       \hline
        Back-lobe gain & & -20 dB\\
       \hline
        Path-loss exponent, $\eta_s$ & 3.76 & 2.5\\
       \hline
        Inter-cell distance & & $1.2\times R_0$\\
       \hline
       AoA error $\sim \mathcal{N}(0, \sigma_{\rm AoA}^2)$ &\multicolumn{2}{c}{$2\degree$}\\
       \hline
    \end{tabular}
    \end{threeparttable}}
\end{table}

In our simulations, we consider  $K_0=15$ UEs randomly located in a bi-dimensional region, under the coverage of $N_s=3$ SBSs working at mmWave frequencies with a carrier frequency of 28 GHz, and one MBS communicating at 2 GHz\textcolor{black}{, however, our solution can be leveraged for applications using different technologies such as WiFi or LiFi.}. Also, we consider three types of service corresponding to an average data rate demand $\overline{D}_j \in\{5, 200, 1500\}$ Mbps. We assume that the traffic request of a UE $j$ is a random variable, which follows a Poisson distribution with intensity $\overline{D}_j = \mathbb{E}\left[D_j(t)\right]$.
Additional simulation parameters can be founded in Table \ref{simu-params}.\\
\noindent
\textbf{Learning parameters.} Since all UEs share the same policy network, $\mathcal{A}$ coincides with the action space. However, a UE $j$ can only be associated with BSs in $\mathcal{A}_j(t) \subseteq \mathcal{A}$. Accordingly, unauthorized actions or connection requests $a_j(t)\not\in\mathcal{A}_j(t)$ are redirected towards the MBS, i.e., they appear as connection requests to the MBS. We fixed the size of the encoding functions $n=128$. All encoding functions are composed of only one hidden multi-layer perceptron (MLP) of $n$ neurons. Both actor and critic comprise also one MLP with $2n$ neurons. All layers use a rectifier linear unit activation. We set the learning rate $\mu$ to $10^{-4}$ and the discounting factor $\gamma=0.6$. Unless specified, we empirically fix the clipping factors to $\epsilon_1=0.01$, $\epsilon_2=0.5$, the time horizon to $T_e=250$ and the UE dropout probability to $p_0=0.95$. Also, we limit UE $j$'s neighborhood $\mathcal{N}_j$ to its $k$-nearest neighbors, where $k\leq15$.\\
\noindent
\textbf{Benchmarks.} As a comparison, we consider the same benchmarks as in \cite{sana2020UA}, i.e., the Max-SNR algorithm, which associates UEs on the basis of links with the maximum SNR, and the centralized heuristic algorithm, which consists in associating UEs, starting from the links with the maximum SNR, and in an iterative way as long as it increases the network utility. Originally proposed in \cite{zhou}, the centralized heuristic algorithm is shown to exhibit good performance, specifically in interference-limited network. Therefore, we use it as a baseline solution in lieu of the optimal solution, infeasible here, due to the network size. However, we recompute the heuristic algorithm every time the network changes, which is also cumbersome. To assess the convergence performance of the proposed algorithm, we define ${r_d(t) = \overline{R}^{\mathrm{Trans.~ RL}}(t) - \overline{R}^{\mathrm{Heur.}}(t)}$, which corresponds to the difference of the average reward over an episode reached by the proposed algorithm compared to the centralized heuristic approach. 
Also, we represent on the histograms, the average performance over 500 random deployments of UEs. 

\subsection{Impact of the hysteretic clipping factors on convergence}
\begin{figure}[!t]
    \centering
    \includegraphics[scale=0.82]{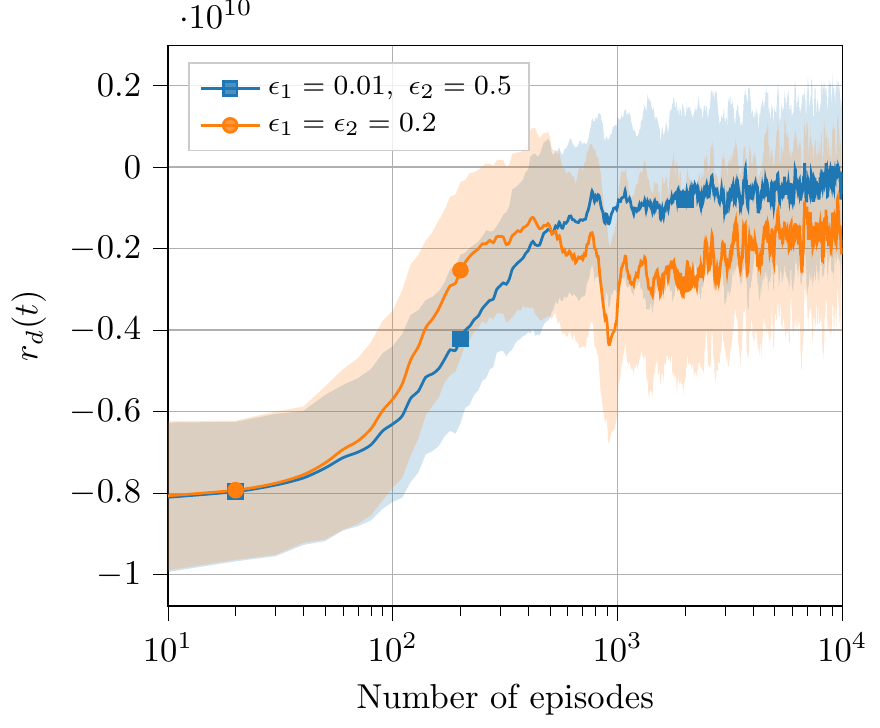}
    \caption{Effect of the hysteretic clipping factors on the convergence. $\alpha=0$ and $D_j(t)=\infty, ~\forall j$. Averaged on a 100-sized rolling window.}
    \label{fig:hppo}
\end{figure}

Here, we evaluate the impact of clipping factors $\epsilon_1$ and $\epsilon_2$ on the convergence. \fig{fig:hppo} shows the evolution of $r_d(t)$ in two settings: $\epsilon_1=\epsilon_2=0.2$, corresponding to the setting of the vanilla PPO proposed in \cite{schulman2017ppo}, and our empirically optimized hysteretic setting $\epsilon_1=0.01,~ \epsilon_2=0.5$. We show that by simply introducing a hysteretic effect in the clipping factors, we notably improve the stability and the learning performance, reaching the same performance as the heuristic algorithm (as $r_d(t)$ converges on average to zero).



\subsection{Policy Transferability Property: Zero shot generalization}

\begin{figure}[!t]
    \centering
    \includegraphics[scale=0.76]{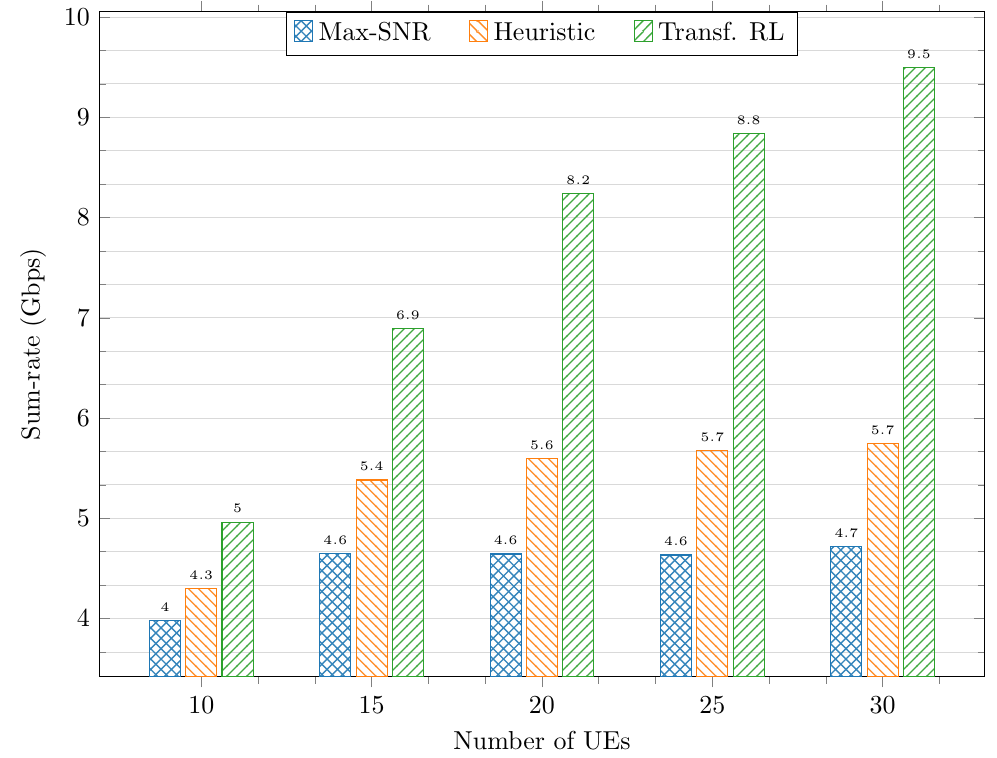}
    \caption{Transferability for case $\alpha=0$ and with network traffic. The PNA is initially trained for 15 UEs.}
    \label{fig:zeroShotTraffic}
\end{figure}
\begin{figure}[!t]
    \centering
    \includegraphics[scale=0.76]{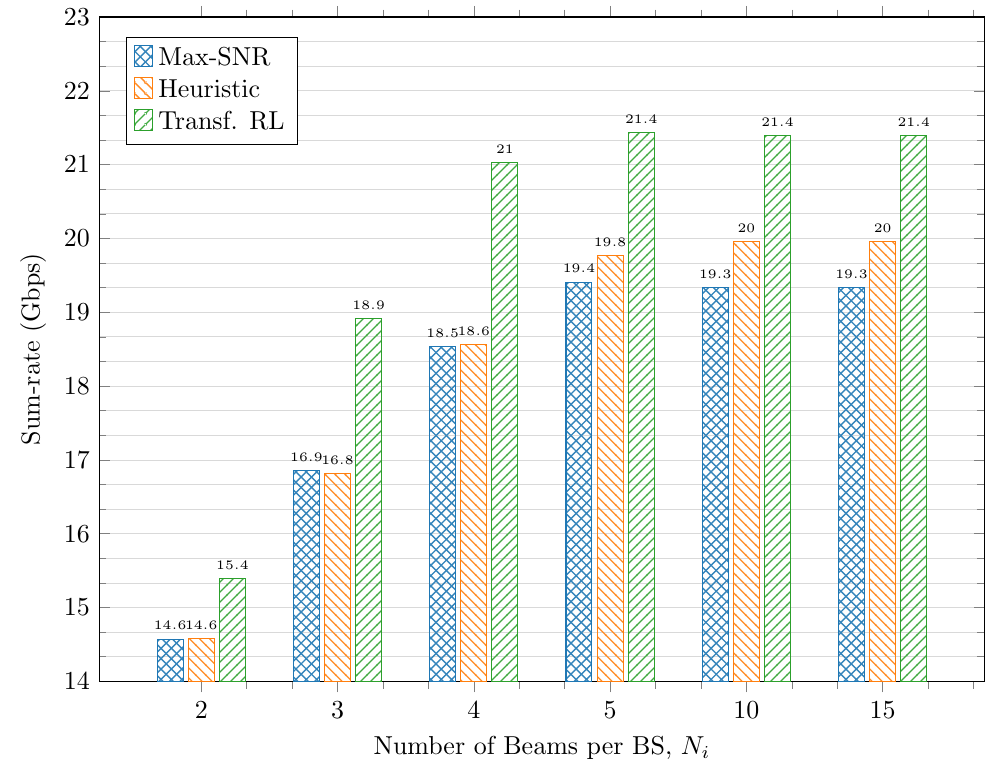}
    \caption{Generalizability for case $\alpha=1$ (no traffic). Here we fix $K=15$ and vary the number of beams $N_i$.}
    \label{fig:zeroShotNi}
\end{figure}
To assess how transferable the proposed algorithm is, we train the PNA for a reference number of users, $K_0=15$ and $N_i=3, ~\forall i$. Then we evaluate on \figs{fig:zeroShotTraffic}{fig:zeroShotNi}, the performance of the trained model for different network deployments with a variable number of UEs $K\in\{10, 20, 25, 30\}$, including changes in the UEs' position and traffic dynamic. In \fig{fig:zeroShotTraffic}, we observe that when we consider the network traffic, the proposed transferable solution clearly outperforms the two benchmarks. Even when the number of UEs doubles from $K_0=15$ to $K=30$, Our solution yields $102.1\%$, $66.66\%$ network sum-rate increase \wrt the max-SNR and heuristic algorithms, respectively. Moreover, an additional feature of the proposed architecture, is that even when the number of beams available per BS later changes (\wrt initial training point, fixed to $N_i=3$), which impacts the collision events, the algorithm still adapts to maintain the system's performance. Indeed, in \fig{fig:zeroShotNi} where we evaluate the performance of the algorithms for different \textcolor{black}{$N_i\in\{2, 3, 4, 5, 10, 15\}$}, we can observe that as $N_i$ increases, implying less and less collisions since $K$ is fixed, the algorithm keeps outperforming the two benchmarks. When $N_i$ becomes greater than $5$, i.e., $\sum_{i=1}^{3} N_i > K=15$, there is no improvement in the sum-rate as there are enough beams to serve all UEs. 

\section{Conclusions and Perspectives}

In this work, we investigated the problem of transferability of user association policies for 5G and beyond networks. We come out with a novel policy architecture and a learning mechanism that enable users to cooperatively learn a robust and transferable user association policy. Our proposed solution exploits neural attention  and deep multi-agent reinforcement learning mechanisms, where agents leverage local and, if available, 
global observations to optimize network utility functions. We achieve transferability. Indeed, with our solution, a policy learned in a given scenario can be transferred 
with zero-shot generalization capability, i.e. without any additional training. Our numerical results showed that the proposed transferable solution provides large gains, indeed doubling the network sum-rate compared to state-of-the-art approaches. The observed benefit is due to the transferability feature, and by jointly considering network traffic and radio channels dynamic during optimization. 

In future work, results of this study will be exploited for applications in Multi-access Edge Computing. 

\bibliographystyle{ieeetr}
\bibliography{biblio}

\end{document}